# GPT has become financially literate: Insights from financial literacy tests of GPT and a preliminary test of how people use it as a source of advice

**Paweł Niszczota*, Sami Abbas**

Poznań University of Economics and Business
Institute of International Business and Economics
Humans & AI Laboratory (HAI Lab)

* Corresponding author: Paweł Niszczota, Poznań University of Economics and Business, al. Niepodległości 10, 61-875 Poznań, Poland, pawel.niszczota@ue.poznan.pl

**Data**

Data and the pre-registration document are available at: https://osf.io/tnbya/

**Declarations of interest**

None.

**Ethical approval**

Informed consent was obtained from all participants and the experiment adhered to appropriate guidelines. The study was approved by an Ethics Committee.

**Financial support**

This research was supported by grant 2021/42/E/HS4/00289 from the National Science Centre, Poland, and grant 004/RID/2018/19 from the Regional Initiative for Excellence program of the Minister of Science and Higher Education of Poland (2019-2022).

**Author statement**

**Paweł Niszczota:** Conceptualization; Methodology; Software; Validation; Formal analysis; Investigation; Resources; Data Curation; Writing – Original Draft; Writing – Review & Editing; Visualization; Supervision; Project administration; Funding acquisition

**Sami Abbas**: Conceptualization; Formal analysis; Investigation; Resources; Data Curation; Writing – Review & Editing.

**Published in *Finance Research Letters* – this is the Author Accepted Manuscript**

For final version (version of record) see:
Niszczota, P., & Abbas, S. (2023). GPT has become financially literate: Insights from financial literacy tests of GPT and a preliminary test of how people use it as a source of advice. *Finance Research Letters*, *58*, 104333. https://doi.org/10.1016/j.frl.2023.104333

**GPT has become financially literate: Insights from financial literacy tests of GPT and a preliminary test of how people use it as a source of advice**

**Abstract**

We assess the ability of GPT–a large language model–to serve as a financial robo-advisor for the masses, by using a financial literacy test. Davinci and ChatGPT based on GPT-3.5 score 66% and 65% on the financial literacy test, respectively, compared to a baseline of 33%. However, ChatGPT based on GPT-4 achieves a near-perfect 99% score, pointing to financial literacy becoming an emergent ability of state-of-the-art models. We use the Judge-Advisor System and a savings dilemma to illustrate how researchers might assess advice-utilization from large language models. We also present a number of directions for future research.

## 1. Introduction

The introduction of ChatGPT – a conversational variant of the third iteration of the Generative Pre-Trained Transformer (GPT) model (Brown et al., 2020) – in November 2022 has caused an enormous increase in public interest in large language models (LLMs, e.g., Roose, 2022; Sample, 2023). Soon after, a number of papers investigated GPT-3.5's performance in various non-trivial tasks (e.g., the Bar exam, which has since become a yardstick to assess progress in LLMs; OpenAI, 2023). While assessments of GPT-3.5 point to mediocre quantitative reasoning abilities (Bommarito et al., 2023), GPT-4 has been shown to make enormous advancements in a wide range of tasks (Bubeck et al., 2023), consistent with the emergence of certain skills (Wei et al., 2022).

Our paper has five contributions. First, we measure how GPT performs on tests aimed at measuring financial literacy. We consider this as a basic test of GPT's ability to provide free or low-cost financial advice to the public. In essence, GPT could be considered a cheap robo-advisor for the masses (e.g., D'Acunto et al., 2019). However, it remains unclear how well GPT will perform on these tasks, given that financial advice usually has a strong numerical component and this is the Achilles heel of LLMs, even the most recent ones (Floridi and Chiriatti, 2020; Srivastava et al., 2022). Second, we compare the actual financial literacy scores with laypeople's predicted scores. This is intended to serve as an early indication of how expected performance diverges from real performance, which could promote overreliance on such technologies. Third, we use a household financial problem (concerning savings) to assess the degree that people utilize advice coming from GPT in a hypothetical scenario. Fourth, we compare the performance of GPT-3.5 in two variants (the conventional variant (*text-davinci-003*; Davinci) and the publicly-available, conversational variant (ChatGPT OpenAI, 2022)) and GPT-4. Finally, we provide a number of directions for future research on LLMs.

There are still little studies on GPT's utility in finance and economics. Notable examples include work published by Dowling and Lucey (2023) and Korinek (2023), who assess whether ChatGPT can be used to aid financial and economic research. Work by Lopez-Lira and Tang (2023) point to a substantial improvement in GPT's ability to predict returns.

## 2. Assessment of financial literacy of GPT-3.5 and GPT-4

### 2.1. Methodology

To measure financial literacy, we used the 'Big Five' items (Mitchell and Lusardi, 2022), and items from the Financial Literacy Baseline Survey (Heinberg et al., 2014), altogether 21 single-choice test questions. These had 2–5 options after the removal of options that indicated a lack of knowledge of the answer (*"Don't know", "Refuse to answer"*), which ensured a more accurate assessment of the actual performance of GPT against an agent who would randomly answer all options, but only in a way that could be potentially correct. An agent guessing would correctly answer 20%–50% of the time, and overall, guessing would lead to an expected score of 33%.

We tested the financial literacy of GPT by asking it each item 20 times, to factor-in the built-in tendency of large language models to provide varied responses. Tests for Davinci were conducted on the most deterministic setting for the sampling temperature (0), as recommended by OpenAI when eliciting factual answers. Additionally, to assess how well GPT performs depending on whether it has been pre-prompted before the actual question to act as a financial advisor, we test both the performance with and without such role-playing. In cases where GPT did not provide the letter indicating the correct answer (*a-e*) but answered correctly, we manually marked the answer as correct.

### 2.2. Results

In **Table 1** we present the mean accuracy of both investigated variants of GPT-3.5 and GPT-4 on financial literacy. Using questions with the absence of a pre-prompt as a benchmark, GPT-3.5 Davinci achieved a 66% financial literacy score, while GPT-3.5 ChatGPT achieved a score of 65%. Interestingly, pre-prompting Davinci to "play" the role of a financial advisor did not improve financial literacy scores, but in fact, reduced them by 6 pp. For ChatGPT there was also a reduction in the score, albeit it was much less pronounced (2 pp). This, in turn, suggests greater sensitivity to input (a greater variance of outcomes) for the more advanced (Davinci) model. The accuracy of GPT does not appear to decrease for items that contain numerical data or longer items (see Appendix).

GPT-4, however, obtained a near-perfect score of 99.3% (without the pre-prompt) and 97.4% (with a pre-prompt). Put differently, GPT-4 exhibits financial literacy: a basic, at the very least, grasp of financial matters. We conducted back-testing, which suggest that this finding is robust to changes in descriptions of the questions or the order of the answers (see Appendix).

**Table 1. Performance of GPT on the financial literacy test (in %)**

| | GPT-3.5 Davinci | | | GPT-3.5 ChatGPT | | | GPT-4 ChatGPT | | |
|---|---|---|---|---|---|---|---|---|---|
| | Without pre-prompt | With pre-prompt | Overall | Without pre-prompt | With pre-prompt | Overall | Without pre-prompt | With pre-prompt | Overall |
| Big Five (5 items) | 100.00 | 100.00 | 100.00 | 63.64 | 60.40 | 62.00 | 100.00 | 100.00 | 100.00 |
| Compound Interest (4 items) | 75.00 | 50.00 | 62.50 | 67.50 | 70.00 | 68.75 | 96.25 | 86.25 | 91.25 |
| Tax-favored assets (5 items) | 60.00 | 60.00 | 60.00 | 57.00 | 65.00 | 61.00 | 100.00 | 100.00 | 100.00 |
| Inflation (2 items) | 50.00 | 50.00 | 50.00 | 50.00 | 10.00 | 30.00 | 100.00 | 100.00 | 100.00 |
| Employer Match (3 items) | 33.33 | 0.00 | 16.67 | 66.67 | 66.67 | 66.67 | 100.00 | 100.00 | 100.00 |
| Risk diversification (2 items) | 0.00 | 0.00 | 0.00 | 100.00 | 100.00 | 100.00 | 100.00 | 100.00 | 100.00 |
| **Financial literacy (all items)** | **66.07** | **60.12** | **63.10** | **65.39** | **63.18** | **64.29** | **99.29** | **97.39** | **98.34** |

*Notes:* The pre-prompt was "*You are a financial advisor.*". Means computed based on 20 trials.

## 3. Advice-utilization from GPT-3.5

To assess laypeople's expectations of GPT performance and assess how knowledge about actual performance impacts advice-utilization, we used the Judge-Advisor System (Sniezek and Buckley, 1995; van Swol and Sniezek, 2005). While the Judge-Advisor System has been used in the past to assess the utilization of algorithmic advice (Logg et al., 2019), it has not yet – to the best of our knowledge – been used to assess advice-taking from GPT.

We tested two hypotheses concerning advice-taking from GPT. First, we expected that predicted scores on the financial literacy test will predict advice-taking behavior on a more specific problem, that would require actual computations. Second, we hypothesized that for people with lower subjective financial knowledge, advice-utilization will be greater.

The study was conducted on GPT-3.5, prior to the release of GPT-4. Note, however, that GPT-4 is not freely available, and findings obtained on the GPT-3.5 variant will be more applicable to alternative large language models such as LLaMA.

### 3.1. Methodology

In the advice-taking task participants rated Davinci. First, they saw 19 financial literacy test items (two were omitted) and were asked how well they thought GPT performed on this test. All participants were then given a dilemma concerning the appropriate monthly payment necessary to reach a goal, which had a single numeric solution. A description of this dilemma and the accuracy of participants' and GPT's answers can be found in the **Appendix**.

Participants provided their answer and later received the mean solution provided by GPT from 20 prompts.

The advice-utilization we used was a weight of advice index (Harvey and Fischer, 1997), defined as:

$$WOA = \frac{final\ answer - initial\ answer}{advice - initial\ answer} \quad (1)$$

WOA typically ranges between 0 – indicating that the judge has entirely disregarded the advice and 1 – indicating that the judge used precisely the answer indicated by the advisor. Values outside of this range were winsorized, in line with previous research on the utilization of advice from machines (Logg et al., 2019).

Details on participants and their recruitment can be found in the Appendix.

The study was pre-registered at https://aspredicted.org/ge4au.pdf and approved by an Ethics Committee. Data and materials are available at https://osf.io/tnbya/.

## 3.2. Results

### 3.2.1. Predicted performance of GPT

Participants predicted that GPT will have a score of 79.32% (*SD* = 17.65%), which was significantly higher than the actual score ($t = 16.47$, $p < .001$). The distribution of predicted scores broken-down based on the median subjective financial knowledge is shown in **Fig. 1**. People with low and high subjective financial knowledge predicted similar performance from GPT ($M_{low} = 77.2\%$ vs $M_{high} = 80.8\%$, $t = 1.32$, $p = .19$).

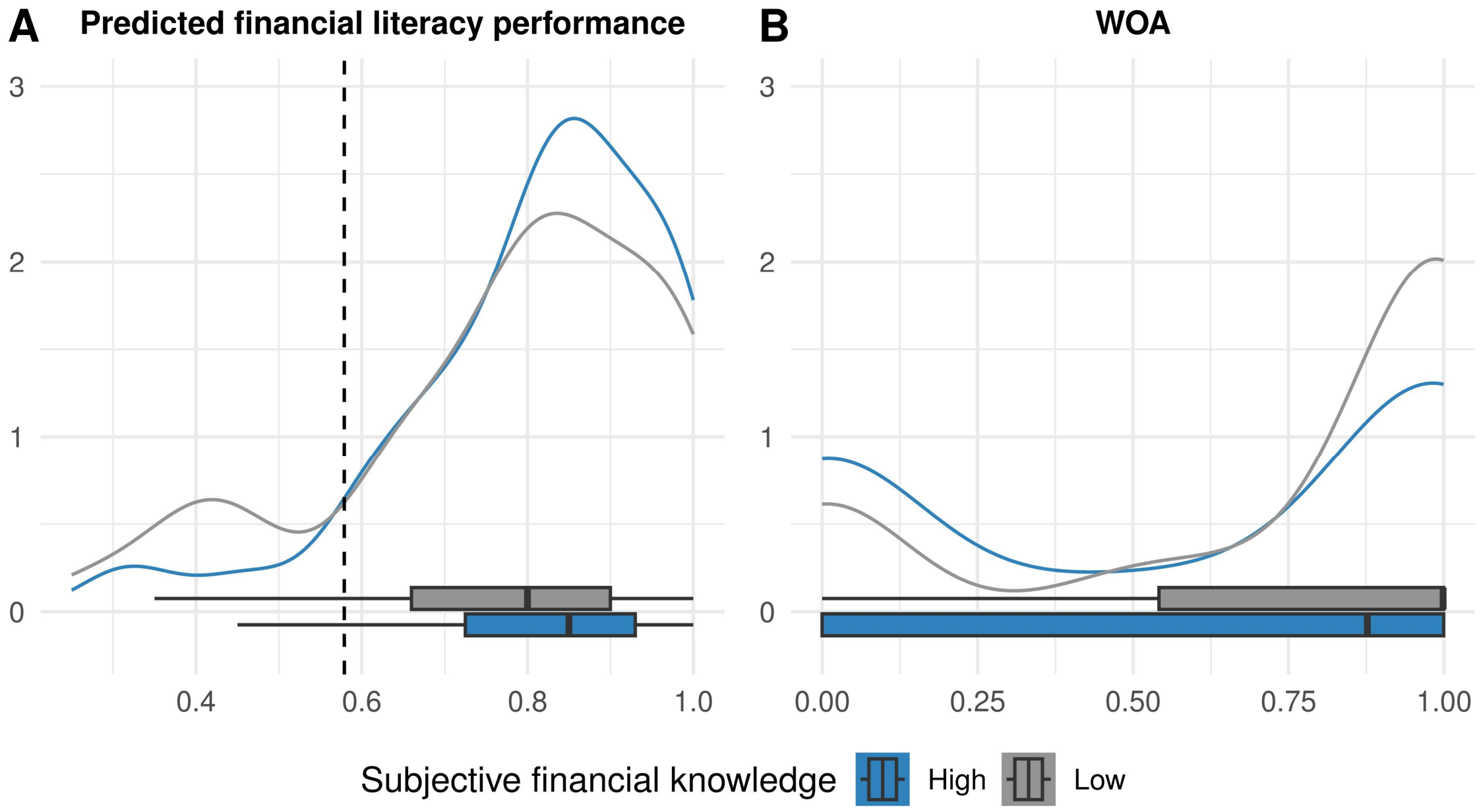

**Fig. 1. The predicted financial literacy performance (A) and advice-utilization (B) of GPT**
*Notes:* This figure shows density plots and box-whisker plots for high and low subjective financial knowledge (split based on the median). The dashed line corresponds to the actual performance of Davinci on the 21-item financial literacy test without pre-prompts. Advice-utilization is measured via WOA.

### 3.2.2. Advice-utilization

The *WOA* score – after performing the pre-registered winsorization – was 0.650 (*SD* = 0.438). This indicated that participants relied more on GPT's estimates than their own. For reference, this is substantially higher than the *WOA* = 0.39 obtained in a recent meta-analysis (Bailey et al., 2022). As illustrated by **Fig. 1**, people with low subjective financial knowledge relied more heavily on advice from GPT than people with high subjective financial knowledge ($M_{low} = 0.740$ vs $M_{high} = 0.588$, $t = 2.39$, $p = .018$).

In **Table 2**, we show regressions with WOA as the dependent variable. As alternative estimation methods, we used OLS and beta regression, the latter being suitable for data in the (0, 1) range, but also computable in the [0, 1] range after a transformation (Smithson and Verkuilen, 2006).

**Table 2. Predictors of advice-utilization from GPT (*WOA*)**

| | OLS | Beta regression |
|---|---|---|
| (Intercept) | 1.16 *** (0.38) | 1.59 (1.20) |
| Predicted score on financial literacy | -0.32 * (0.19) | -0.66 (0.60) |
| Subjective financial knowledge | -0.06 ** (0.02) | -0.15 ** (0.07) |
| Objective financial knowledge | 0.02 (0.03) | 0.07 (0.10) |
| Prior interaction with GPT | -0.14 * (0.08) | -0.35 (0.22) |
| Age | -0.03 (0.10) | -0.12 (0.31) |
| Gender (baseline = *female*) | Yes | Yes |
| *N* | 184 | 184 |
| $R^2$ adjusted | 0.035 | 0.078 |

*Notes:* Robust standard errors are shown in parentheses. For the beta regression, the table reports coefficients for the mean equation. * $p < 0.1$ ** $p < 0.05$ *** $p < 0.01$

Regressions do not support the hypothesis positing that higher expected scores on the financial literacy tests lead to greater utilization of GPT's advice (see Appendix for marginal effects). However, subjective financial knowledge seems to show a robust link to advice-taking: for both estimation methods, people with lower subjective financial knowledge relied more heavily on advice from GPT.

## 4. Discussion

Our analysis shows an enormous improvement in the financial literacy of LLMs. GPT-3.5 possess limited financial literacy, obtaining a 65-66% score against a 33% score that would be expected from random responses, consistent with a weakness in LLMs in quantitative reasoning (Bommarito et al., 2023; Floridi and Chiriatti, 2020; Srivastava et al., 2022). However, ChatGPT based on the GPT-4 model obtained a near-perfect score of 99%. Given that these were tests designed to study laypeople, it is reasonable to posit that some of the *most advanced* LLMs now possess the capability to serve as robo-advisors (Chak et al., 2022; D'Acunto et al., 2019) for the masses.

Our study also provides preliminary evidence on how much people might rely on information from GPT concerning financial matters. Using a simple savings problem and the Judge-Advisor System, we estimate advice-utilization to be substantial (*WOA* = 0.650), and even higher in people with low subjective financial knowledge (*WOA* = 0.740). The latter remain at a greater risk of overreliance on large language models. However, it is not yet known to what extent laypeople will use GPT for financial advice: some earlier research suggests that people with less financial knowledge might be less likely to use robo-advice (Isaia and Oggero, 2022; Niszczota and Kaszás, 2020).

## 4.1. The future of large language models in finance

### 4.1.1. Utility of financial advice from LLMs

On balance, LLMs should provide good financial advice *most* of the time, with room for further improvement as time passes. Performance can further be improved after fine-tuning by certified professionals, e.g., investment fund managers, CFAs, or PhDs in finance. For example, GPT recommends using a cash windfall to pay off debt and establish an emergency fund (along with other recommendations), and to pay off high-interest debt (e.g., credit card debt) first. Many people will recognize that LLMs can serve as a free, preliminary source of advice, and may only consult with a human financial advisor for high-stakes decisions.

It is difficult to precisely assess the benefits of this free, unbiased, and mostly-accurate source of advice. Assuming that financial advice serves as a substitute for financial literacy (see, e.g., Disney et al., 2015), advanced LLMs can become a useful source of financial advice for people with lower financial literacy. Research has shown that the less financially-literate more often rely on informal (less accurate) sources of advice, such as their friends and family (van Rooij et al., 2011). Access to sophisticated LLMs can potentially produce a plethora of benefits for them. For example, van Rooij et al. (2011) show that people in the highest financial literacy quartile are nearly six-times more likely to invest on the stock market, which can serve as an upper bound for the effect of adherence to good advice from LLMs (for this specific outcome). See the Appendix for other possible effects that obtaining access to a credible source of financial advice might bring to the less financially-literate.

### 4.1.2. Risk of overreliance

LLMs are not eager to acknowledge incompetence. In fact, LLMs such as GPT are known to "hallucinate" answers, i.e. answer prompts counterfactually, but deliver it in a confident fashion that would appear to be correct to a layperson. Although more recent iterations of these models are less prone to such hallucinations (Ouyang et al., 2022), it remains a significant weakness of these models. Overconfidence of ChatGPT would be of course paradoxical, as advice should not contain the same flaws that it is meant to address (Lewis, 2018).

Research on advice-taking suggests why the overconfidence of LLMs is problematic. Price and Stone (2004) have shown that people may conflate confidence with competence in an advisor, i.e. they show a "confidence heuristic". Assuming that an analogous relationship exists for AI advisors, people will incorrectly think that an LLM is correct if the model delivers the incorrect answer confidently. Models that are more conservative in their responses (e.g., GPT-4) might not be freely available, requiring either a monthly subscription fee or offered with a considerable premium when accessed via an API.

## 4.2. Limitations

### 4.2.1. Large language models

There are several limitations to our work. First, we used proprietary language models from one source (OpenAI). Since the introduction of ChatGPT, several companies have developed alternative models (e.g., Bard or LLaMA). Second, responses from LLMs are sensitive to prompting, and perhaps more accurate responses could be achieved using different prompts. Third, even though GPT has vastly improved its financial literacy, it is not evident that it will retain this literacy. Some research already points to the possibility of GPT-4 losing some of its abilities (Chen et al., 2023).

### 4.2.2. Advice utilization

The Judge-Advisor System is clearly just a simplified method illustrating advice-taking, and does not map into many features inherent in financial decision-making. First of all, financial advice from professionals takes into account the characteristics of the advisee such as risk preferences, age, income etc. Secondly, financial advice should take into consideration national regulations, concerning, for example, taxation: advice that might be sound in one country may not be sound in another country due to taxation differences.

### 4.2.3. Other limitations

We present some further limitations in the Appendix.

## 4.3. Future directions

For the benefit of future research, in **Fig. 2** we summarize the key directions that the research on financial advice might go. Future research could attempt to answer how financially literate other LLMs are. Researchers could also look how the circumstances of the decision (e.g., the country of residence of the advisee) or characteristics of the advisee (e.g., risk tolerance or wealth) will affect the quality of advice provided by LLMs. Finally, a wide-range of outcomes could be investigated to assess the effectiveness of advice from LLMs.

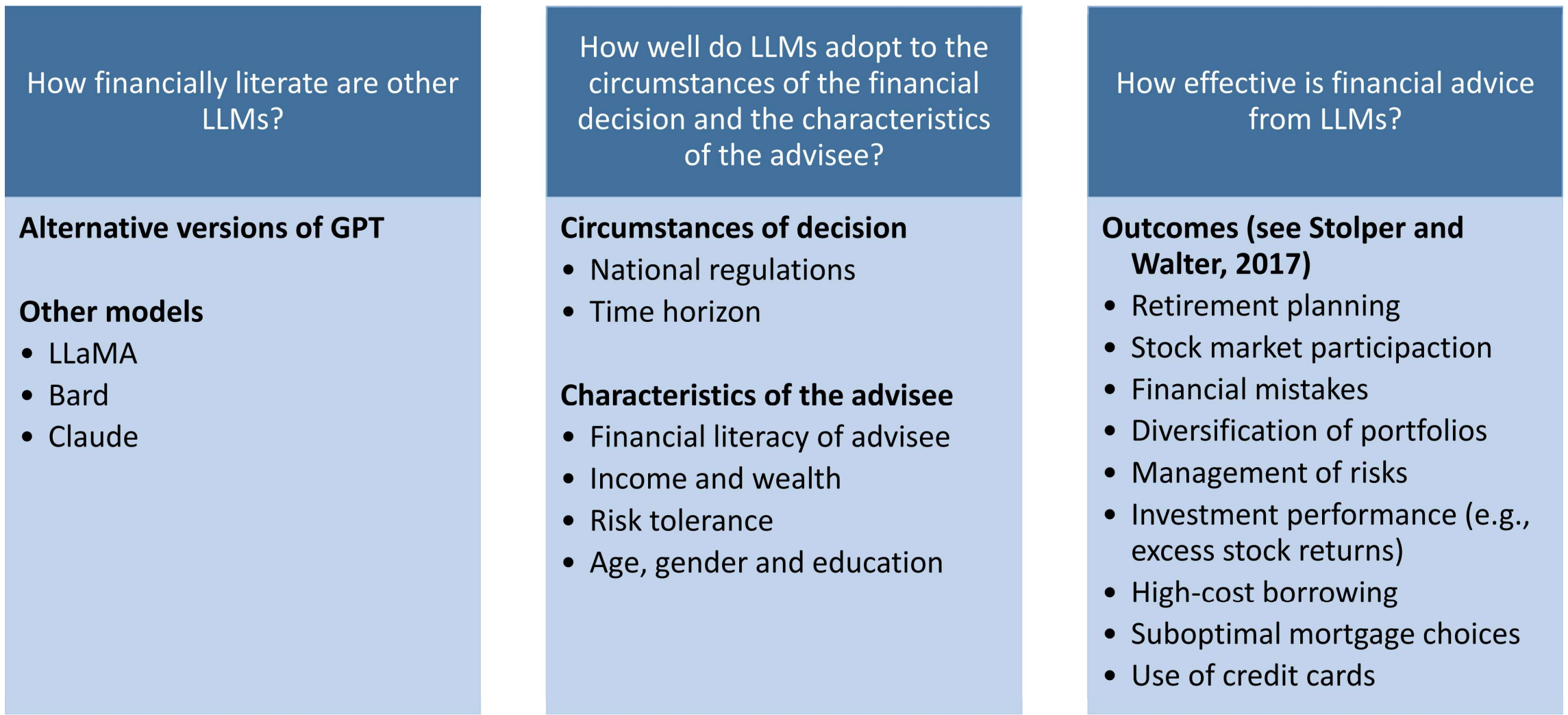

**Fig. 2. Future directions for research on financial advice from LLMs**

# Appendix

## Details on participants in advice utilization task

To assess advice utilization from GPT, we recruited 200 participants from Prolific (Palan and Schitter, 2018), that had a 98% or higher approval rating, were located and born in the US, and whose first language was English. As preregistered, 16 participants that didn't correctly answer both attention check questions were excluded, leaving a final sample size of 184. The mean age of participants was 47.1 years (*SD* = 13.8); 32% of them interacted with GPT in the past. We used one item to assess their subjective financial knowledge *("My knowledge concerning finance is good."*, rated on a scale of 1 (*fully disagree*) to 7 (*fully agree*), with a mean score of 3.86 (*SD* = 1.50). We also applied a six-item financial knowledge test to assess objective financial (investment) knowledge (*M* = 4.69, *SD* = 1.15).

Participants received a 0.80 GBP flat fee for participation. To incentivize participants to put effort into the task, we informed them that the top 10% of participants will receive a bonus (0.50 GBP).

## Materials

### General introduction

This survey concerns Generative Pre-trained Transformers (GPT) - sophisticated natural language processing (AI) models. These models are better known in the form of ChatGPT, a conversational variant, that is available to the public since November 2022. This study concerns an alternative (non-conversational) version of GPT from the same company (OpenAI), called Davinci (text-davinci-003). We will refer to Davinci simply as GPT from now on. GPT uses complex machine learning (or deep learning) techniques, and was trained on an enormous amount of data. It uses 175 billion parameters in its neural network, making it able to understand human language and generate human-like text. If a user asks a question about a particular topic, GPT is able to provide a relevant response based on its understanding of the context (e.g., finance). This survey consists of two main parts: You will be asked to estimate how well GPT will do on a financial literacy test, consisting of 19 items. You will see a short financial problem, which we would like you to answer.

### Introduction to Part I

Let's start with Part I. You will see 19 items from a financial literacy test on the page, and on the page after that, you will be asked to estimate how well you think GPT would do on this financial literacy test.

### Predicted financial literacy of GPT

Below are the 19 questions we gave GPT in the financial literacy test. Please note that you don't have to read them very carefully one by one -- they are only meant to give you an understanding of the nature of the questions that we asked GPT.

1. Suppose you had $100 in a savings account and the interest rate was 2% per year. After 5 years, how much do you think you would have in the account if you left the money to grow? a) More than $102 b) Exactly $102 c) Less than $102
2. Imagine that the interest rate on your savings account was 1% per year and inflation was 2% per year. After 1 year, how much would you be able to buy with the money in this account? a) More than today b) Exactly the same c) Less than today
3. If interest rates rise, what will typically happen to bond prices? a) They will rise b) They will fall c) They will stay the same d) There is no relationship between bond prices and the interest rate
4. Suppose you put 1,000 dollars in an account that earns 5% interest per year, every year. You never invest additional money and you never withdraw money or interest payment. So in the first year, you earn 50 dollars in interest. In year 4, how much will this account earn? a) Less than 50 dollars b) 50 dollars c) More than 50 dollars
5. Suppose you invest $2,500 and earn 7% per year on this investment. How many years will it take for your total investment to be worth $5,000? a) Between 0 and 5 years b) Between 5 and 15 years c) Between 15 and 45 years d) More than 45 years
6. Consider the following scenario: Jack and Jill are twins. At the age of 20, Jack started contributing $20 a month to a savings account. After 20 years, at the age of 40, he

stopped adding to his savings, but he left the money in the account. Jill didn't start to save until she was 40. Then, she saved $20 a month until she retired 20 years later at the age of 60. Suppose both Jack and Jill earned 6% interest per year on their savings. When they both retired at age 60, who had more money? a) Jack b) Jill c) They had the same amount

7. Pam is deciding between 2 options; Option A: invest $1,000 in a certificate of deposit that earns 5% interest. Pam would not add or remove any money from this investment for the next 30 years. Option B: Invest $1,000 in a savings account that earns 5% interest. Move the interest earned on this account every year into a safe at home. Pam would not add or remove any other money from the savings account or the safe for the next 30 years. At the end of 30 years, which of these options would provide the most money? a) Option A b) Option B
8. Suppose that by the year 2020 your income has doubled and prices of all goods have doubled too. In 2020, how much will you be able to buy with your 2020 income? a) More than today b) The same amount as today c) Less than today
9. Rita must choose between two job offers. She wants to select a job with a salary that will afford her a higher standard of living for the next few years. Job A offers a 3% raise every year, while Job B will not provide a raise for the next few years. If Rita chooses Job A, she will live in City A. If Rita chooses Job B, she will live in City B. Rita finds that the price of goods and services today are about the same in both areas. Prices are expected to rise, however, by 4% in City A every year, and stay the same in City B. Based on her concerns about standard of living, what should Rita do? a) Take Job A b) Take Job B c) Take either one: she will be able to afford the same future standard of living in both places
10. Which of the following is an accurate statement about investment returns? a) Usually, investing $5,000 in shares of a single company is safer than investing $5,000 in a fund which invests in shares of many companies in multiple industries. b) Usually, investing $5,000 in shares of a single company is less safe than investing $5,000 in a fund which invests in shares of many companies in different industries. c) Usually, investing $5,000 in shares of a single company is less safe than investing $5,000 in a fund which invests in shares of many companies in different industries.
11. Suppose you are a member of a stock investment club. This year, the club has about $200,000 to invest in stocks and the members prefer not to take a lot of risk. Which of the following strategies would you recommend to your fellow members? a) Put all of the money in one stock b) Put all of the money in two stocks c) Put all of the money in a stock indexed fund that tracks the behavior of 500 large firms in the United States
12. When you invest in an employer's retirement savings plan such as a 401(k), your contributions are taxed: a) Either before you invest them or when you withdraw them during retirement, but not both times. b) Both before you invest them and when you withdraw them during retirement. c) Once a year on or before April 15. d) When you reach age 65.
13. Both Irene and her employer contribute every year to her employer:sponsored 401(k) plan. Irene has worked at the company for twenty years, and is fully vested in her plan. Suppose Irene leaves her job or gets fired. Which of the following statements is true? a) If she is no longer working for the company, the whole plan balance is forfeited, because her benefits are tied to her job. b) If she gets fired, the company has the right to decide how much of her total plan balance she will get. c) If she voluntarily leaves her job, she forfeits all of her employer's contributions. d) Even if she leaves her job or gets fired, she is still entitled to the entire plan balance.

14. Which of the following statements are true? a) In any type of IRA or 401(k) account, all of the money in your account grows tax free. b) If you have a traditional IRA or 401(k), you make contributions out of pretax income and pay income tax at your future tax rate when you withdraw the funds. c) Both are true.
15. This year, Marge's salary is $100,000 and she contributes $10,000 of her salary to a traditional 401(k) offered by her employer. Her current tax rate is 28% . In 40 years, when Marge retires, the money will have grown to $160,000. Her tax rate during retirement will fall to 20% . Which of the following is true? a) This year, Marge should pay income taxes on her entire salary. During retirement, she will pay 20% tax on whatever she withdraws from her plan. b) This year, Marge should pay income taxes on only $90,000. During retirement, she will pay the same deferred 28% tax rate on whatever she withdraws from her plan. c) This year, Marge should pay income taxes on only $90,000. During retirement, she will pay 20% tax on whatever she withdraws from her plan. d) This year, Marge should pay income taxes on only $90,000. During retirement, she will pay no tax on whatever she withdraws from her plan.
16. Which of the following is a true statement? a) You will lose money that you personally invested in your 401(k) if you switch jobs. b) You will be charged income tax as well as tax on dividends and increases in the value of your stock if you invest through a 401(k). c) Unless you are undergoing significant hardship, you cannot withdraw money from a 401(k) without penalty until you reach a certain age. d) All of the above
17. Alice wants to invest $1,000 for retirement this year. Her new employer will fully match her 401(k) contributions, up to $10,000 per year. All else being equal, which of the following options will give Alice the highest total amount at the end of the year? a) Alice contributes $1,000 to her 401(k) plan and invests that money in mutual fund A. At the end of the year, mutual fund A earned a 5% return. b) Alice does not contribute to her 401(k) plan but she invests $1,000 in mutual fund B outside of her 401(k) plan. At the end of the year, mutual fund B earned a 20% return. c) Alice does not contribute to her 401(k) plan, but she invests $1,000 in mutual fund A outside of her 401(k) plan. At the end of the year, mutual fund A earned a 5% return.
18. David's new job offers a 401(k). His employer provides a 50% match up to 2,000. How much should David invest at least in order to obtain the maximum amount of money from the employer match? a) $0 b) $500 c) $1,000 d) $2,000 e) $4,000
19. You have decided to set aside 15% of your salary for retirement. You work at a firm where your employer matches your contribution to the 401(k) plan, dollar by dollar, up to 5% of your salary. Which of these statements is correct? a) If you contribute up to 5% of your salary, the employer match is equivalent to a 100% return on your contribution. b) What the employer contributes should not play any role in your decision. c) It is always a good idea to contribute less than what the employer contributes.

How well do you think GPT scored on this test in percent (from 0 to 100)?

**___ %**

Note that in this particular multiple choice test, a person randomly guessing would expect a score of 31%

**Introduction to Part II**

Now, let's move on to Part II. You will see a short financial problem.

**Financial problem**

Now, imagine the following financial problem. Mary made an agreement with her older brother that she will be able to buy his sailing boat four years from now for $33,333, in cash. She wants to pay in cash, so she wants to make monthly deposits into a savings account, that can earn 4.5% APR. How much should Mary deposit to her savings account to be able to save $33,333 that she will need four years from now to buy the sailing boat (assuming no taxes, and that the savings rate will be constant through the entire time)?

Please write your answer somewhere (e.g., on a piece of paper) and then enter it in the box below:

**$ ___**

**Advisor estimate**

We have asked GPT 20 times the same financial problem that you just saw and answered. The average answer GPT provided in these 20 attempts was: $ 637.50

**Final answer**

Knowing your own estimate of the financial problem, and knowing how GPT answered the financial problem in 20 attempts ($637.50), please provide your final answer to the financial problem.

Also, please note 10% of participants with the most accurate final answers will receive a bonus of 0.50 GBP.

My final answer is:

**$___**

**Introduction to final part of survey**

Thank you for completing Part I and II. Now you will answer some final questions on a numbers of pages, after which we will provide you with the survey code.

**Attention checks**

What was the information you received after you made your first estimate about the monthly payment? [4 options, 1 was correct]

In the second attention check, participants answered a question concerning an image (i.e., completed a task that is easy for a human, and difficult for a machine)

**Demographics**

What is your gender? [male/female/non-binary/prefer not to say]

What is your age (in years)?

To what extent do you disagree or agree with the following statement below? Please rate on a scale of 1 (*fully disagree*) to 7 (*fully agree*): My financial knowledge is good.

**GPT knowledge**

Have you heard about GPT or ChatGPT (not including this survey)? [Yes/No]

How many of the studies that you participated on Prolific in the past concerned (or explicitly used) ChatGPT or GPT? [No studies at all (apart from this one)/One study/Two to five studies/More than five studies]

Have you interacted with GPT or ChatGPT? [Yes/No]

**Objective financial (investment) knowledge**

**Investment knowledge test items.** Items 1-4 were originally used in van Rooij et al. (2011). Items 5-6 were originally used in Agnew and Szykman (2005). The items were presented in random order. The minimum score in the investment knowledge test was 0 and the maximum was 6.

1. Considering a long time period (e.g., 10 or 20 years), which asset normally gives the highest return: savings accounts, bonds or stocks?
2. Normally, which asset displays the highest fluctuations over time: savings accounts, bonds, or stocks?
3. Stocks are normally riskier than bonds - is this statement True or False?
4. When an investor spreads money among different unrelated assets, does the risk of losing money: increase, decrease or stay the same?
5. If you were to invest $1000 in a stock fund, would it be possible to have less than $1000 when you decide to withdraw or move it to another fund?
6. High yield bond funds are invested in bonds with strong credit ratings - is this statement True or False?

**Comments**

If you have any comments, please put them in the box below: [box]

## Supplementary tables

**Table A1. Items used to test financial literacy**

| Item code | Subtype | Prompt | Expected score (from chance) | Correct answers without preprompt (%) | | | Correct answers with preprompt (%) | | |
|---|---|---|---|---|---|---|---|---|---|
| | | | | GPT-3.5 Davinci | GPT-3.5 ChatGPT | GPT-4 ChatGPT | GPT-3.5 Davinci | GPT-3.5 ChatGPT | GPT-4 ChatGPT |
| Big5_1 | Big 5 | Suppose you had $100 in a savings account and the interest rate was 2% per year. After 5 years, how much do you think you would have in the account if you left the money to grow? a) More than $102 b) Exactly $102 c) Less than $102 | 33% | 60% | 20% | 100% | 50% | 0% | 100% |
| Big5_2 | Big 5 | Imagine that the interest rate on your savings account was 1% per year and inflation was 2% per year. After 1 year, how much would you be able to buy with the money in this account? a) More than today b) Exactly the same c) Less than today | 33% | 100% | 100% | 100% | 100% | 100% | 100% |
| Big5_3 | Big 5 | Do you think that the following statement is true or false? “Buying a single company stock usually provides a safer return than a stock mutual fund.” a) True b) False | 50% | 100% | 100% | 100% | 100% | 100% | 100% |

| Item code | Subtype | Prompt | Expected score (from chance) | Correct answers without preprompt (%) | | | Correct answers with preprompt (%) | | |
|---|---|---|---|---|---|---|---|---|---|
| | | | | GPT-3.5 Davinci | GPT-3.5 ChatGPT | GPT-4 ChatGPT | GPT-3.5 Davinci | GPT-3.5 ChatGPT | GPT-4 ChatGPT |
| Big5_4 | Big 5 | If interest rates rise, what will typically happen to bond prices? a) They will rise b) They will fall c) They will stay the same d) There is no relationship between bond prices and the interest rate | 25% | 100% | 100% | 100% | 100% | 100% | 100% |
| Big5_5 | Big 5 | True or false? A 15-year mortgage typically requires higher monthly payments than a 30-year mortgage, but the total interest paid over the life of the loan will be less. a) True b) False | 50% | 100% | 100% | 100% | 100% | 100% | 100% |
| FLBS_CI2 | FLBS Compound Interest | Suppose you put 1,000 dollars in an account that earns 5% interest per year, every year. You never invest additional money and you never withdraw money or interest payment. So in the first year, you earn 50 dollars in interest. In year 4, how much will this account earn? a) Less than 50 dollars b) 50 dollars c) More than 50 dollars | 33% | 100% | 100% | 100% | 100% | 100% | 100% |
| FLBS_CI3 | FLBS Compound Interest | Suppose you invest $2,500 and earn 7% per year on this investment. How many years will it take for your total investment to | 25% | 50% | 100% | 65% | 43% | 85% | 45% |

| Item code | Subtype | Prompt | Expected score (from chance) | Correct answers without preprompt (%) | | | Correct answers with preprompt (%) | | |
|---|---|---|---|---|---|---|---|---|---|
| | | | | GPT-3.5 Davinci | GPT-3.5 ChatGPT | GPT-4 ChatGPT | GPT-3.5 Davinci | GPT-3.5 ChatGPT | GPT-4 ChatGPT |
| | | be worth $5,000? a) Between 0 and 5 years b) Between 5 and 15 years c) Between 15 and 45 years d) More than 45 years | | | | | | | |
| FLBS_CI4 | FLBS Compound Interest | Consider the following scenario: Jack and Jill are twins. At the age of 20, Jack started contributing $20 a month to a savings account. After 20 years, at the age of 40, he stopped adding to his savings, but he left the money in the account. Jill didn’t start to save until she was 40. Then, she saved $20 a month until she retired 20 years later at the age of 60. Suppose both Jack and Jill earned 6% interest per year on their savings. When they both retired at age 60, who had more money? a) Jack b) Jill c) They had the same amount | 33% | 53% | 5% | 100% | 0% | 0% | 100% |

| Item code | Subtype | Prompt | Expected score (from chance) | Correct answers without preprompt (%) | | | Correct answers with preprompt (%) | | |
|---|---|---|---|---|---|---|---|---|---|
| | | | | GPT-3.5 Davinci | GPT-3.5 ChatGPT | GPT-4 ChatGPT | GPT-3.5 Davinci | GPT-3.5 ChatGPT | GPT-4 ChatGPT |
| FLBS_CI5 | FLBS Compound Interest | Pam is deciding between 2 options; Option A: invest $1,000 in a certificate of deposit that earns 5% interest. Pam would not add or remove any money from this investment for the next 30 years. Option B: Invest $1,000 in a savings account that earns 5% interest. Move the interest earned on this account every year into a safe at home. Pam would not add or remove any other money from the savings account or the safe for the next 30 years. At the end of 30 years, which of these options would provide the most money? a) Option A b) Option B | 50% | 83% | 65% | 100% | 98% | 95% | 100% |
| FLBS_I2 | FLBS Inflation | Suppose that by the year 2020 your income has doubled and prices of all goods have doubled too. In 2020, how much will you be able to buy with your 2020 income? a) More than today b) The same amount as today c) Less than today | 33% | 100% | 100% | 100% | 55% | 10% | 100% |

| Item code | Subtype | Prompt | Expected score (from chance) | Correct answers without preprompt (%) | | | Correct answers with preprompt (%) | | |
|---|---|---|---|---|---|---|---|---|---|
| | | | | GPT-3.5 Davinci | GPT-3.5 ChatGPT | GPT-4 ChatGPT | GPT-3.5 Davinci | GPT-3.5 ChatGPT | GPT-4 ChatGPT |
| FLBS_I3 | FLBS Inflation | Rita must choose between two job offers. She wants to select a job with a salary that will afford her a higher standard of living for the next few years. Job A offers a 3% raise every year, while Job B will not provide a raise for the next few years. If Rita chooses Job A, she will live in City A. If Rita chooses Job B, she will live in City B. Rita finds that the price of goods and services today are about the same in both areas. Prices are expected to rise, however, by 4% in City A every year, and stay the same in City B. Based on her concerns about standard of living, what should Rita do? a) Take Job A b) Take Job B c) Take either one: she will be able to afford the same future standard of living in both places | 33% | 0% | 0% | 100% | 5% | 10% | 100% |

| Item code | Subtype | Prompt | Expected score (from chance) | Correct answers without preprompt (%) | | | Correct answers with preprompt (%) | | |
|---|---|---|---|---|---|---|---|---|---|
| | | | | GPT-3.5 Davinci | GPT-3.5 ChatGPT | GPT-4 ChatGPT | GPT-3.5 Davinci | GPT-3.5 ChatGPT | GPT-4 ChatGPT |
| RD3 | FLBS Risk diversification | Which of the following is an accurate statement about investment returns? a) Usually, investing $5,000 in shares of a single company is safer than investing $5,000 in a fund which invests in shares of many companies in multiple industries. b) Usually, investing $5,000 in shares of a single company is less safe than investing $5,000 in a fund which invests in shares of many companies in different industries. c) Usually, investing $5,000 in shares of a single company is less safe than investing $5,000 in a fund which invests in shares of many companies in different industries. | 33% | 50% | 100% | 100% | 50% | 100% | 100% |

| Item code | Subtype | Prompt | Expected score (from chance) | Correct answers without preprompt (%) | | | Correct answers with preprompt (%) | | |
|---|---|---|---|---|---|---|---|---|---|
| | | | | GPT-3.5 Davinci | GPT-3.5 ChatGPT | GPT-4 ChatGPT | GPT-3.5 Davinci | GPT-3.5 ChatGPT | GPT-4 ChatGPT |
| RD4 | FLBS Risk diversification | Suppose you are a member of a stock investment club. This year, the club has about $200,000 to invest in stocks and the members prefer not to take a lot of risk. Which of the following strategies would you recommend to your fellow members? a) Put all of the money in one stock b) Put all of the money in two stocks c) Put all of the money in a stock indexed fund that tracks the behavior of 500 large firms in the United States | 33% | 50% | 100% | 100% | 50% | 100% | 100% |
| TF2 | FLBS Tax-favored assets | When you invest in an employer's retirement savings plan such as a 401(k), your contributions are taxed: a) Either before you invest them or when you withdraw them during retirement, but not both times. b) Both before you invest them and when you withdraw them during retirement. c) Once a year on or before April 15. d) When you reach age 65. | 25% | 85% | 70% | 100% | 98% | 95% | 100% |

| Item code | Subtype | Prompt | Expected score (from chance) | Correct answers without preprompt (%) | | | Correct answers with preprompt (%) | | |
|---|---|---|---|---|---|---|---|---|---|
| | | | | GPT-3.5 Davinci | GPT-3.5 ChatGPT | GPT-4 ChatGPT | GPT-3.5 Davinci | GPT-3.5 ChatGPT | GPT-4 ChatGPT |
| TF3 | FLBS Tax-favored assets | Both Irene and her employer contribute every year to her employer:sponsored 401(k) plan. Irene has worked at the company for twenty years, and is fully vested in her plan. Suppose Irene leaves her job or gets fired. Which of the following statements is true? a) If she is no longer working for the company, the whole plan balance is forfeited, because her benefits are tied to her job. b) If she gets fired, the company has the right to decide how much of her total plan balance she will get. c) If she voluntarily leaves her job, she forfeits all of her employer's contributions. d) Even if she leaves her job or gets fired, she is still entitled to the entire plan balance. | 25% | 100% | 100% | 100% | 100% | 100% | 100% |

| Item code | Subtype | Prompt | Expected score (from chance) | Correct answers without preprompt (%) | | | Correct answers with preprompt (%) | | |
|---|---|---|---|---|---|---|---|---|---|
| | | | | GPT-3.5 Davinci | GPT-3.5 ChatGPT | GPT-4 ChatGPT | GPT-3.5 Davinci | GPT-3.5 ChatGPT | GPT-4 ChatGPT |
| TF4 | FLBS Tax-favored assets | Which of the following statements are true? a) In any type of IRA or 401(k) account, all of the money in your account grows tax free. b) If you have a traditional IRA or 401(k), you make contributions out of pretax income and pay income tax at your future tax rate when you withdraw the funds. c) Both are true. | 33% | 8% | 15% | 100% | 15% | 30% | 100% |

| Item code | Subtype | Prompt | Expected score (from chance) | Correct answers without preprompt (%) | | | Correct answers with preprompt (%) | | |
|---|---|---|---|---|---|---|---|---|---|
| | | | | GPT-3.5 Davinci | GPT-3.5 ChatGPT | GPT-4 ChatGPT | GPT-3.5 Davinci | GPT-3.5 ChatGPT | GPT-4 ChatGPT |
| TF5 | FLBS Tax-favored assets | This year, Marge's salary is $100,000 and she contributes $10,000 of her salary to a traditional 401(k) offered by her employer. Her current tax rate is 28% . In 40 years, when Marge retires, the money will have grown to $160,000. Her tax rate during retirement will fall to 20% . Which of the following is true? a) This year, Marge should pay income taxes on her entire salary. During retirement, she will pay 20% tax on whatever she withdraws from her plan. b) This year, Marge should pay income taxes on only $90,000. During retirement, she will pay the same deferred 28% tax rate on whatever she withdraws from her plan. c) This year, Marge should pay income taxes on only $90,000. During retirement, she will pay 20% tax on whatever she withdraws from her plan. d) This year, Marge should pay income taxes on only $90,000. During retirement, she | 25% | 100% | 100% | 100% | 100% | 100% | 100% |

| Item code | Subtype | Prompt | Expected score (from chance) | Correct answers without preprompt (%) | | | Correct answers with preprompt (%) | | |
|---|---|---|---|---|---|---|---|---|---|
| | | | | GPT-3.5 Davinci | GPT-3.5 ChatGPT | GPT-4 ChatGPT | GPT-3.5 Davinci | GPT-3.5 ChatGPT | GPT-4 ChatGPT |
| | | will pay no tax on whatever she withdraws from her plan. | | | | | | | |
| TF6 | FLBS Tax-favored assets | Which of the following is a true statement? a) You will lose money that you personally invested in your 401(k) if you switch jobs. b) You will be charged income tax as well as tax on dividends and increases in the value of your stock if you invest through a 401(k). c) Unless you are undergoing significant hardship, you cannot withdraw money from a 401(k) | 25% | 0% | 0% | 100% | 0% | 0% | 100% |

| Item code | Subtype | Prompt | Expected score (from chance) | Correct answers without preprompt (%) | | | Correct answers with preprompt (%) | | |
|---|---|---|---|---|---|---|---|---|---|
| | | | | GPT-3.5 Davinci | GPT-3.5 ChatGPT | GPT-4 ChatGPT | GPT-3.5 Davinci | GPT-3.5 ChatGPT | GPT-4 ChatGPT |
| | | without penalty until you reach a certain age. d) All of the above | | | | | | | |
| EM2 | FLBS Employer Match | Alice wants to invest $1,000 for retirement this year. Her new employer will fully match her 401(k) contributions, up to $10,000 per year. All else being equal, which of the following options will give Alice the highest total amount at the end of the year? a) Alice contributes $1,000 to her 401(k) plan and invests that money in mutual fund A. At the end of the year, mutual fund A earned a 5% return. b) Alice does not contribute to her 401(k) plan but she invests $1,000 in mutual | 33% | 50% | 100% | 100% | 50% | 100% | 100% |

| Item code | Subtype | Prompt | Expected score (from chance) | Correct answers without preprompt (%) | | | Correct answers with preprompt (%) | | |
|---|---|---|---|---|---|---|---|---|---|
| | | | | GPT-3.5 Davinci | GPT-3.5 ChatGPT | GPT-4 ChatGPT | GPT-3.5 Davinci | GPT-3.5 ChatGPT | GPT-4 ChatGPT |
| | | fund B outside of her 401(k) plan. At the end of the year, mutual fund B earned a 20% return. c) Alice does not contribute to her 401(k) plan, but she invests $1,000 in mutual fund A outside of her 401(k) plan. At the end of the year, mutual fund A earned a 5% return. | | | | | | | |
| EM3 | FLBS Employer Match | David's new job offers a 401(k). His employer provides a 50% match up to 2,000. How much should David invest at least in order to obtain the maximum amount of money from the employer match? a) $0 b) $500 c) $1,000 d) $2,000 e) $4,000 | 20% | 0% | 0% | 100% | 0% | 0% | 100% |

| Item code | Subtype | Prompt | Expected score (from chance) | Correct answers without preprompt (%) | | | Correct answers with preprompt (%) | | |
|---|---|---|---|---|---|---|---|---|---|
| | | | | GPT-3.5 Davinci | GPT-3.5 ChatGPT | GPT-4 ChatGPT | GPT-3.5 Davinci | GPT-3.5 ChatGPT | GPT-4 ChatGPT |
| EM4 | FLBS Employer Match | You have decided to set aside 15% of your salary for retirement. You work at a firm where your employer matches your contribution to the 401(k) plan, dollar by dollar, up to 5% of your salary. Which of these statements is correct? a) If you contribute up to 5% of your salary, the employer match is equivalent to a 100% return on your contribution. b) What the employer contributes should not play any role in your decision. c) It is always a good idea to contribute less than what the employer contributes. | 33% | 100% | 100% | 100% | 50% | 100% | 100% |

*Notes*: Financial literacy items are sourced from Heinberg et al. (2014), and Mitchell and Lusardi (2022). GPT-3.5 ChatGPT prompting was conducted on the January 9 (2023) version. GPT-4 ChatGPT prompting was conducted on the July (2023) versions. The advice-utilization task was performed on 19 items, i.e. it excluded items Big5_3 and Big5_5.

**Table A2. Descriptive statistics**

| Variable | Stats / Values | Frequency | Histogram |
|---|---|---|---|
| WOA (non-winsorized) | Mean (SD) : 0.71 (0.95) min ≤ med ≤ max: -0.28 ≤ 0.98 ≤ 11.89 IQR (CV) : 1 (1.32) | 89 distinct values | 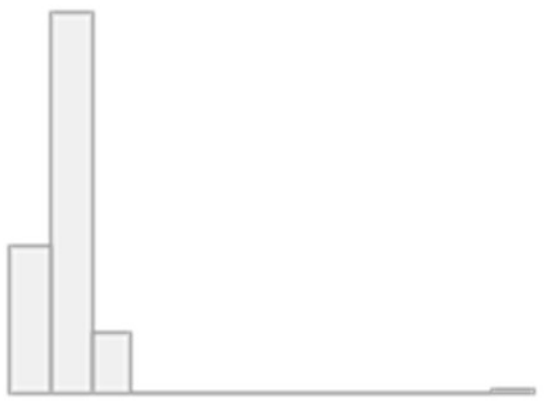 |
| WOA (winsorized) | Mean (SD) : 0.65 (0.44) min ≤ med ≤ max: 0 ≤ 0.98 ≤ 1 IQR (CV) : 1 (0.68) | 63 distinct values | 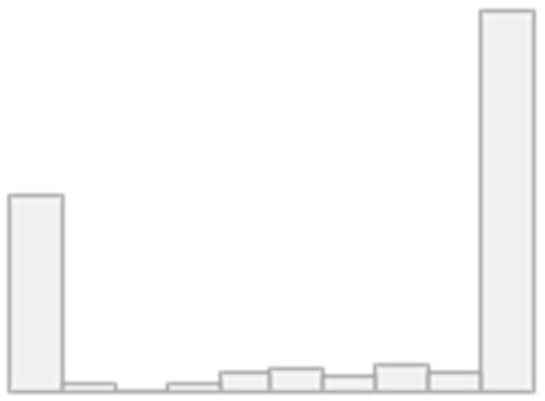 |
| MAPE (pre-advice) | Mean (SD) : 1653.4 (3786.13) min ≤ med ≤ max: 0 ≤ 84.71 ≤ 45561.38 IQR (CV) : 3828.3 (2.29) | 107 distinct values | 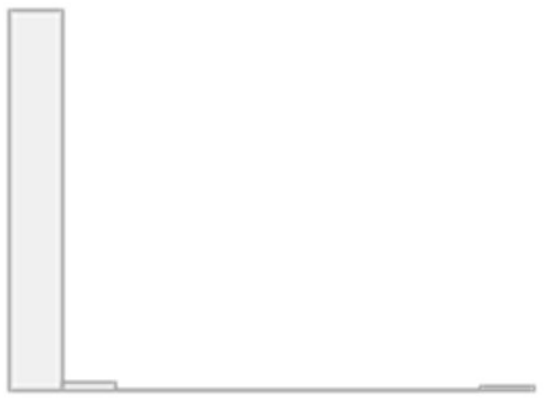 |
| MAPE (post-advice) | Mean (SD) : 509.9 (1336.79) min ≤ med ≤ max: | 86 distinct values | 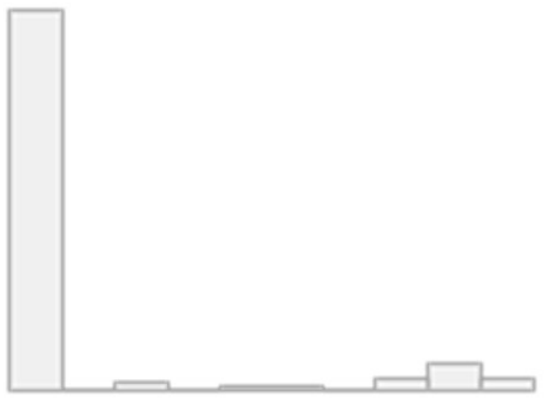 |

| Variable | Stats / Values | Frequency | Histogram |
|---|---|---|---|
| | 0 ≤ 2.34 ≤ 4938.5<br>IQR (CV) : 9.84 (2.62) | | |
| Female participant | Min : 0<br>Mean : 0.54<br>Max : 1 | 0 : 85 (46.2%)<br>1 : 99 (53.8%) | 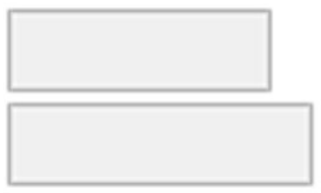 |
| Age | Mean (SD) : 41.71 (13.83)<br>min ≤ med ≤ max:<br>19 ≤ 38 ≤ 81<br>IQR (CV) : 20.5 (0.33) | 56 distinct values | 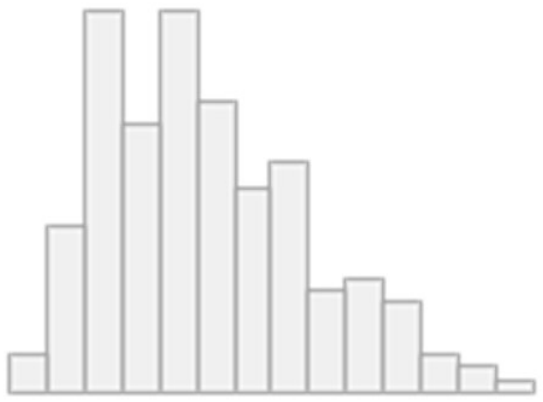 |
| Financial knowledge (objective) | Mean (SD) : 4.69 (1.15)<br>min ≤ med ≤ max:<br>1 ≤ 5 ≤ 6<br>IQR (CV) : 2 (0.24) | 1 : 1 ( 0.5%)<br>2 : 6 ( 3.3%)<br>3 : 22 (12.0%)<br>4 : 46 (25.0%)<br>5 : 54 (29.3%)<br>6 : 55 (29.9%) | 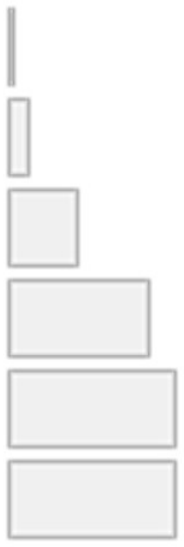 |

| Variable | Stats / Values | Frequency | Histogram |
|---|---|---|---|
| Financial knowledge (subjective) | Mean (SD) : 3.86 (1.5)<br>min ≤ med ≤ max:<br>1 ≤ 4 ≤ 7<br>IQR (CV) : 2 (0.39) | 1 : 14 ( 7.6%)<br>2 : 20 (10.9%)<br>3 : 40 (21.7%)<br>4 : 46 (25.0%)<br>5 : 38 (20.7%)<br>6 : 20 (10.9%)<br>7 : 6 ( 3.3%) | 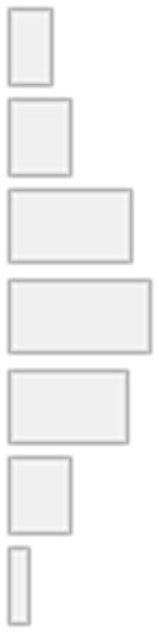 |
| Have you heard about GPT or ChatGPT (not including this survey)? | 1. 0<br>2. 1 | 35 (19.0%)<br>149 (81.0%) | 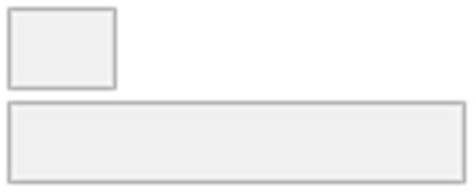 |
| How many of the studies that you participated on Prolific in the past concerned (or explicitly used) ChatGPT or GPT? | Mean (SD) : 0.59 (0.87)<br>min ≤ med ≤ max:<br>0 ≤ 0 ≤ 3<br>IQR (CV) : 1 (1.48) | 0 : 117 (63.6%)<br>1 : 32 (17.4%)<br>2 : 29 (15.8%)<br>3 : 6 ( 3.3%) | 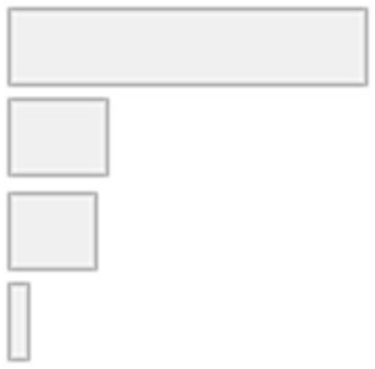 |
| Have you interacted with GPT or ChatGPT? | 1. 0<br>2. 1 | 126 (68.5%)<br>58 (31.5%) | 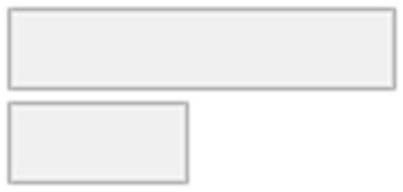 |

**Table A3. Frequency table for Weight of Advice**

| WOA | Winsorized | Frequency | Frequency % |
|---|---|---|---|
| **0** | No | 40 | 21.74% |
| **0** | Yes | 6 | 3.26% |
| **0 < WOA < 1** | No | 66 | 35.87% |
| **1** | No | 52 | 28.26% |
| **1** | Yes | 20 | 10.87% |
| | | **184** | **100.00%** |

**Table A4. Correlations between accuracy of GPT in responses and features of test item**

| | Pre-prompt | Pearson correlations | | | Spearman correlations | | |
|---|---|---|---|---|---|---|---|
| | | Numerical item (1 = yes, 0 = no) | Length of test item | Length of test item (logged) | Numerical item (1 = yes, 0 = no) | Length of test item | Length of test item (logged) |
| Davinci | No | -0.05 | 0.00 | -0.04 | -0.13 | -0.12 | -0.12 |
| | Yes | -0.10 | 0.04 | -0.01 | -0.14 | -0.02 | -0.02 |
| ChatGPT | No | 0.05 | 0.09 | 0.05 | 0.07 | 0.04 | 0.04 |
| | Yes | 0.13 | 0.24 | 0.24 | 0.12 | 0.32 | 0.32 |

*Notes*: $N = 19$. Correlations are computed based on mean accuracy on test item in 20 trials.

**Table A5. Predictors of advice utilization from GPT (*WOA*) – marginal effects**

| | **Beta regression** |
|---|---|
| Predicted score on financial literacy | -0.16<br>(0.14) |
| Subjective financial knowledge | -0.05*<br>(0.03) |
| Objective financial knowledge | 0.02<br>(0.03) |
| Prior interaction with GPT | -0.08<br>(0.05) |
| Age | -0.01<br>(0.05) |
| Gender (baseline = *female*) | Yes |
| *N* | 184 |
| $R^2$ adjusted | 0.078 |

*Notes:* Marginal effects were computed using the *betamfx* function from the *mfx* package in R (Fernihough and Henningsen, 2019). Robust standard errors are shown in parentheses. * p < 0.1 ** p < 0.05 *** p < 0.01.

## The accuracy of GPT-3.5 models

In contrast to financial literacy scores, there were substantial differences between Davinci and ChatGPT on this particular task. We gave this task to Davinci at three different temperatures (the minimal (0), the default (0.7), and the midpoint between these (0.35)) and ChatGPT. To measure accuracy, we used the Mean Absolute Percentage Error (MAPE; Harvey and Fischer, 1997). Results, presented in **Table A6**, suggest that Davinci in the lowest (most deterministic) setting has a MAPE of 0.4, indicating that in 20 trials it deviated 0.4% on average from the correct answer. ChatGPT performed substantially worse, with a MAPE score of 14.8. This, however, was substantially two orders of magnitude lower than the mean pre-advice estimates of participants, whose answers exhibited substantial variation.

**Table A6. Performance of GPT-3.5-based Davinci and ChatGPT on task (20 trials)**

| Model | Temperature | Mean | SD | MAPE |
|---|---|---|---|---|
| Davinci | 0 | 637.50 | 0 | 0.4 |
| Davinci | 0.35 | 639.63 | 42.01 | 3.7 |
| Davinci | 0.70 | 657.27 | 64.56 | 7.8 |
| ChatGPT | - | 571.27 | 138.99 | 14.8 |

*Notes:* MAPE = | (estimate – true) / true | × 100.

## Back-testing

As it is possible that large language models might simply be recalling answers to questions they were trained, on not providing any actual reasoning. Therefore, we performed back-testing by:

1. Changing the description of questions by changing the names of the subjects in the question and changing between personal pronouns and hypothetical test subjects. The contents of the questions were also reorganized to avoid bearing any similarity to the previous iterations of the question to avoid redundancies and structural similarities to the previous iterations of the question.
2. Changing the order of answers (to test whether it is not simply pointing to the indicating correct answer in the financial test)

Tests were conducted on GPT-4. We asked all of the 21 financial literacy items. GPT-4 provided the correct answer in 100% cases (test 1), however when asked again with different ordered answers (test 2) it provided the correct answer in 96% of the cases. This is in line with the original findings, and points to the robustness of our findings.

## Financial literacy categories

To measure financial literacy we used the Big 5 (Heinberg et al., 2014) and a number of items from (Heinberg et al., 2014). Below is a short description of the categories of items from the latter (verbatim), highlighting why it is important to assess them:

**Compound Interest.** Understanding the difference between simple and compound interest, and how quickly interest accumulates can help individuals both appreciate the importance of starting to save early and the dangers of borrowing at very high interest rates.

**Tax-favored assets.** Retirement assets invested in tax advantaged vehicles such as 401(k)s and IRAs benefit from tax exemptions on contributions, capital gains, or withdrawals, allowing for more rapid potential growth.

**Inflation.** Individuals need to understand the potential reduction in purchasing power over time due to inflation in order to assess saving and borrowing decisions in real rather than nominal terms. This is particularly important given the long horizons typical in planning for retirement.

**Employer Match.** Failure to contribute up to the employer's matching threshold is often the equivalent of leaving money on the table.

**Risk diversification.** Individuals should not put all of their eggs in one basket, but rather choose well-diversified portfolios and avoid investing in only one asset, particularly if that asset is their employer's company stock.

### Possible positive effects of access to state-of-the-art large language models

In an excellent literature review, Stolper and Walter (2017) summarize findings from the literature linking financial literacy with financial outcomes, and show that people with higher financial literacy:

1. Are more likely to plan for retirement.
2. Are more likely to invest in the stock market.
3. Commit less financial mistakes.
4. Hold better-diversified portfolios.
5. Better manage their risks.
6. Have higher excess stock returns.
7. Are less likely to have high-cost borrowing.
8. Are less likely to make suboptimal mortgage choices.
9. Are more likely to use their credit cards efficiently.

## Other limitations / suggestions for future financial literacy and advice-utilization studies

### Other financial literacy items and financial dilemmas

We used a specific combination of financial literacy items and only one hypothetical financial dilemma. Alternative items and problems might produce different responses.

### Comparison against professional financial advisors

Future studies can gauge how large language models compare against professionals (human financial advisors). This could include measuring their advanced (and not only basic) financial competencies, and the assessment of how well people adhere to the advice from different sources.

### Making predictions for different categories

We asked participants to only predict the general performance of GPT on the financial literacy test, and not in the specific categories (Compound Interest, Tax-favored assets, Inflation, Employer Match, Risk diversification). Future research might investigate how people will judge the performance of LLMs for different types of questions, and assess how the actual performance of LLMs and human financial advisors deviates from these expectations.

### Framing

It is plausible that the degree of advice-utilization might depend on how the large language model is framed (e.g., we included the number of parameters used in one model). Researchers may assess whether expectations of LLM performance remain similar for different framings

### Giving participants more time to acquaint themselves with the materials

We conducted an online experiment, in which forcing participants to carefully read all pieces of information (e.g., all financially literacy items) would lead to higher costs. New studies could investigate advice utilization in the absence of time constraints.

### Alternative assessment of participants financial literacy

For a better understanding of the link between financial literacy of advice from humans and LLM, future studies might use more detailed assessments of advisees' financial literacy.

### Possibility of disutility

Most studies discuss artificial intelligence as intended to be socially beneficial, and much effort is put into aligning AI with societal goals. However, LLMs are – like any technology – a double-edged sword, with the potential to be used by bad actors. For example, GPT can be asked to provide advice concerning tax evasion. While much effort is put into addressing these issues – either via fine-tuning on a massive scale (Ouyang et al., 2022) – there are ways to overcome them and to use LLMs in a socially-harmful way.

## References for Appendix